\pdfoutput=1

\documentclass[11pt]{article}

\usepackage{acl}

\usepackage{times}
\usepackage{latexsym}

\usepackage[T1]{fontenc}

\usepackage[utf8]{inputenc}

\usepackage{microtype}

\usepackage{inconsolata}

\usepackage{graphicx}

\usepackage{times}
\usepackage{latexsym}
\usepackage{comment}
\usepackage{graphicx}
\usepackage{booktabs}
\usepackage{enumitem}
\usepackage{multirow}
\usepackage{array}  
\usepackage{ragged2e} 
\usepackage{listings}  

\usepackage{longtable}
\usepackage{adjustbox}
\usepackage{tcolorbox} 
\usepackage{lmodern}   
\usepackage{caption}
\usepackage{float}
\usepackage{lscape}

\usepackage{lmodern}
\usepackage{hyperref}
\usepackage[normalem]{ulem}
\usepackage[T1]{fontenc}

\usepackage[utf8]{inputenc}

\usepackage{microtype}

\usepackage{inconsolata}
\newcommand{\model}{\textsc{MHQA}}

\title{\model: A Diverse, Knowledge Intensive Mental Health Question Answering Challenge for Language Models}

 \author{Suraj Racha
 \hspace{1em} Prashant Joshi
 \hspace{1em} Anshika Raman
 \hspace{1em} Nikita Jangid\\  
 {\bf Mridul Sharma} 
 \hspace{1em}  {\bf Ganesh Ramakrishnan}
 \hspace{1em}  {\bf Nirmal Punjabi}\\
        Indian Institute of Technology Bombay\\
         \texttt{\{23d1627, 23m0771, 210050014, nikitajangid, 30005914\}@iitb.ac.in,}\\
         \texttt{ganesh@cse.iitb.ac.in, npunjabi@iitb.ac.in}}

\begin{document}
\maketitle
\begin{abstract}
Mental health remains a challenging problem all over the world, with issues like depression, anxiety becoming increasingly common. Large Language Models (LLMs) have seen a vast application in healthcare, specifically in answering medical questions. However, there is a lack of standard benchmarking datasets for question answering (QA) in mental health. Our work presents a novel multiple choice dataset, MHQA (Mental Health Question Answering), for benchmarking Language models (LMs). Previous mental health datasets have focused primarily on text classification into specific labels or disorders. MHQA, on the other hand, presents question-answering for mental health focused on four key domains: anxiety, depression, trauma, and obsessive/compulsive issues, with diverse question types, namely, factoid, diagnostic, prognostic, and preventive. We use PubMed abstracts as the primary source for QA. 
We develop a rigorous pipeline for LLM-based identification of information from abstracts based on various selection criteria and converting it into QA pairs. Further, valid QA pairs are extracted based on post-hoc validation criteria.
Overall, our MHQA dataset consists of 2,475 expert-verified gold standard instances called MHQA-gold and \textasciitilde56.1k pairs pseudo labeled using external medical references. We report F1 scores on different LLMs along with few-shot and supervised fine-tuning experiments, further discussing the insights for the scores.
\end{abstract}

\section{Introduction}
Recent times have witnessed increased mental health challenges yet it remains an underestimated issue \cite{mh-issue-1, mh-issue-2}.
Annually, mental health issues, including depression, anxiety, and bipolar disorders, affect 30\% of the global population \cite{30_pop}.
Language models (LMs) have shown large potential for medical solutions owing to their advancing reasoning abilities \cite{llm-healthcare}. 
\begin{figure}
    \centering
    \includegraphics[width=1\linewidth]{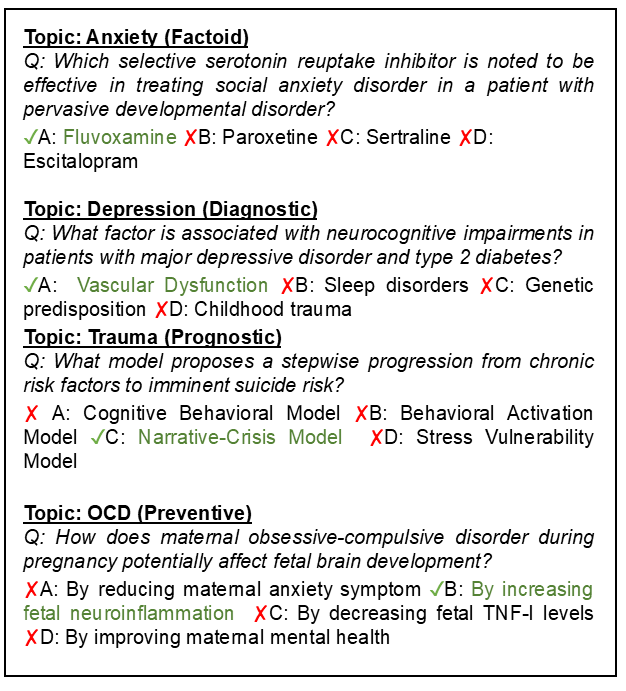}
    \caption{An instance of MHQA dataset: Questions are categorized into four topics, each followed by four answer choices, with one correct option}
    \label{fig:example QA}
\end{figure}

Question answering (QA) is an effective way to assess the ability of any model. While a lot of datasets like MedMCQA \cite{medmcqa}, MedQA \cite{medqa}, BioASQ \cite{bioasq}, PubMedQA \cite{pubmedqa} exist for general medical QA, including both long-form and multiple-choice tasks, they lack targeted questions for mental health. Moreover, limited question types restrict the ability to comprehensively evaluate a language model.
\begin{figure*}
    \centering
    \includegraphics[width=1\linewidth]{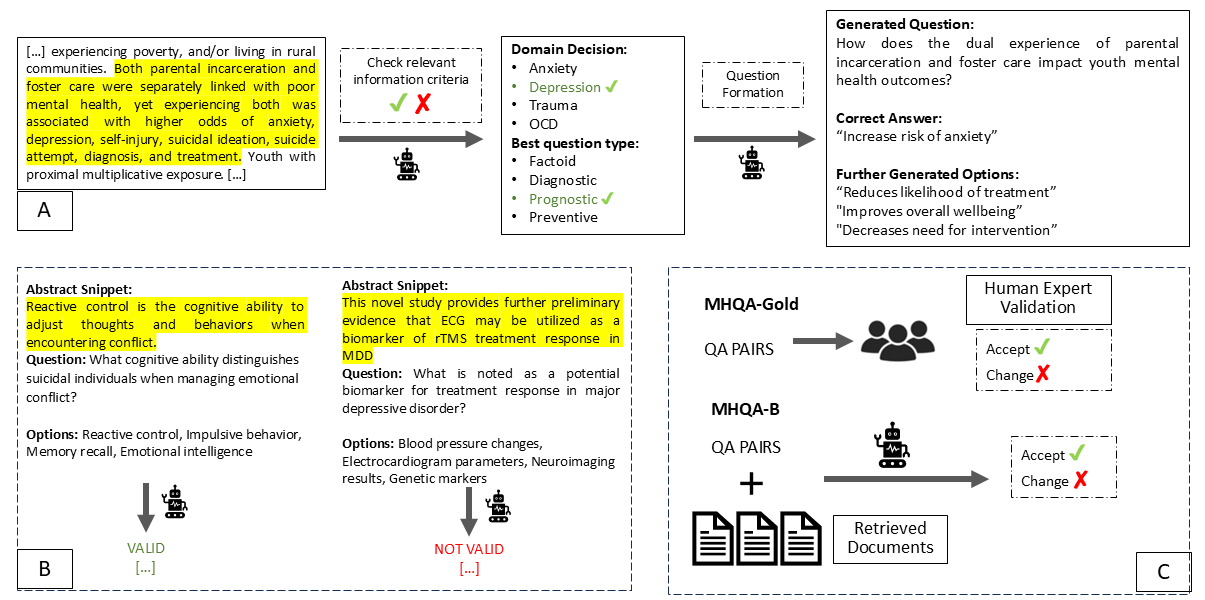}
    \caption{Overall Framework used for dataset generation. (A) Shows selection of abstracts and conversion into questions. (B) A post-hoc validation process to remove inconsistent questions. (C) Human and Pseudo annotation process for correction of inconsistent options.}
    \label{fig:qa-framework}
\end{figure*}
While previous mental health datasets have focused largely on counseling based QA and classification of posts from Reddit and other social platforms \cite{cams-dataset, sad-dataset, angst}, such datasets do not address knowledge-driven question answering. 
Additionally, classification posts may not be clinically or scientifically grounded and often reflect the general audience's opinions. Moreover, the overall data size is limited to just a few thousand, making it difficult to scale.

To this end, we introduce a novel mental health dataset called Mental Health Question Answering (MHQA) for answering mental health-related questions grounded in real-world scenarios and scientific knowledge \footnote{We release the code base and dataset at \url{https://github.com/joshiprashanthd/mhqa}.}. MHQA consists of questions based on abstracts of PubMed research articles on mental health. Unlike unverified content, PubMed articles are peer-reviewed and validated through proper referencing.

MHQA is a large-scale mental health QA dataset designed for comprehensive training and analysis of language models. MHQA dataset has the following unique features: (1) \textasciitilde 58.6k QA pairs, each with four options and a correct answer. (2) A subset of 2,475 QA pairs, manually annotated and verified by three human expert. We refer this as MHQA-Gold and the remaining set as MHQA-B. (3) The QA spans targeted domains with questions based on four aspects of mental health, namely, (i) Anxiety, (ii) Depression, (iii) Trauma, (iv) Obsessive and Compulsive issues. These domains are among the most commonly addressed mental health concerns \cite{Goodwin2015-iq, National_Collaborating_Centre_for_Mental_Health_UK2011-mi}. (4) The questions span a diverse set of tasks, where each question is categorized into factoid, diagnostic, prognostic, and preventive. 
This approach enhances the dataset by including diverse tasks for reasoning and factual evaluation. 
Figure \ref{fig:example QA} shows examples of typical QA pairs picked from different domain. 

We develop a robust pipeline to convert general knowledge evidence from the given abstracts using the GPT-4o-mini model into QA pairs through various criteria and post-hoc verification methodology.
We use MHQA-Gold to benchmark performance on various language models including Llama-3/3.1 8B, GPT-3.5, and GPT-4o along with different discriminative models 
We show experiments with few-shot chain of thought (CoT) prompting strategy for LLMs and supervised finetuning (SFT) on discriminative models using MHQA-B dataset.
Further, we discuss the performance of various models on individual domains and question types.
MHQA can encourage researchers to further develop and test NLP models for mental health QA focusing on reasoning and knowledge comprehension. 

We summarize our contributions as follows: (i) A novel multiple-choice dataset MHQA based on specific information collected from PubMed abstracts. The dataset contains \textasciitilde58.6k QA instances spanning four broad mental health categories, with various question types requiring strong domain knowledge and reasoning ability. (ii) A subset of 2,475 instances from the original dataset with human-annotated answers serves as the gold dataset. (iii) Benchmarked scores of MHQA-Gold on various language models across settings, including zero-shot, few-shot CoT, and SFT.

\section{Related Works}
\subsection{Mental Health Related Datasets}
Social media platforms like Reddit and their posts have been extensively used for building mental health corpus for classification. These include CLPsyc \cite{clpsyc} for classifying into disorder or causes and ANGST \cite{angst} for comorbid anxiety and depression classification. Dreaddit and Depression Reddit focus on single label depression classification. 
SAD \cite{sad-dataset} and CAMS \cite{cams-dataset} on the other hand consider for depression classification into various degrees. DATD considers a Reddit post for anxiety.
The recent dataset called IMHI \cite{mentallama} consists of many previously presented classification datasets to form a unified mental health benchmark aimed at evaluating model performances.
Several datasets and methods also support the development of mental health dialogue systems. ConvCounsel \cite{chen2024convcounselconversationaldatasetstudent} focuses on active listening in student counseling, while MentalQA \cite{alhuzali2024mentalqaannotatedarabiccorpus} provides Arabic mental health QA data. SMILECHAT \cite{qiu2024smilesingleturnmultiturninclusive}creates multi-turn dialogues from single-turn QA using ChatGPT. ESConv \cite{liu2021emotionalsupportdialogsystems} and AugESC \cite{zheng2023augescdialogueaugmentationlarge} offer emotional support dialogues, with 1,053 and 102k dialogues, respectively. Methods like SMILE generate multi-turn conversations using LLMs, addressing data scarcity, while PsyEval \cite{jin2024psyevalsuitementalhealth} assesses LLMs’ mental health knowledge, diagnostic ability, and emotional support skills. General medical QA datasets like TREC QA \cite{trecqa}, BioASQ 2019 \cite{bioasq}, PubMedQA \cite{pubmedqa} also include questions on mental health, even though not explicitly classified. MedMCQA \cite{medmcqa} has a separate category for psychiatry with \textasciitilde4.1k QA pairs.

\subsection{Mental Health Models}
Many discriminative language models have been trained extensively on mental health corpuses or dataset. Mental-BERT \cite{MentalBERT}, Mental-RoBERTa, Mental-XLNet \cite{mentalXLNET}, Mental-LongFormer \cite{mentalXLNET}, etc. have been trained on Reddit mental health posts, thereby improving the inherent understanding of the specific domain. Recently \cite{mentallama} introduced Mental-Llama, which is the first kind of work for LLM-based training on mental health datasets. Specifically, Llama-2 models were extensively fine-tuned on the IMHI dataset for classification and reasoning-based outputs. Similar works also include Mental-Flan-T5 and Mental-Alpaca. Recent work by \cite{interpret-mental-health} has also shown an analysis of the performance of various LLMs like GPT on general mental health domain understanding, including classification and condition identification.




\begin{table}[t]
\centering
\resizebox{0.90\columnwidth}{!}{%
\begin{tabular}{@{}llrr@{}}
\toprule
\textbf{Category} & \textbf{Type} & \textbf{MHQA-Gold} & \textbf{MHQA-B} \\ \midrule
\multirow{4}{*}{Anxiety} & Factoid & 85 & 2,250 \\ 
 & Diagnostic & 171 & 4,363 \\ 
 & Prognostic & 136 & 3,367 \\ 
 & Preventive & \textbf{212} & \textbf{5,284} \\ \cmidrule(l){2-4} 
 & \textbf{Total} & 604 & 15,264 \\ \midrule
\multirow{4}{*}{Depression} & Factoid & 90 & 3826 \\ 
 & Diagnostic & \textbf{207} & \textbf{8,315} \\ 
 & Prognostic & 152 & 6,252 \\ 
 & Preventive & 168 & 7,183 \\ \cmidrule(l){2-4} 
 & \textbf{Total} & 616 & 25,576 \\ \midrule
\multirow{4}{*}{Trauma} & Factoid & 26 & 498 \\ 
 & Diagnostic & \textbf{242} & \textbf{3,572} \\ 
 & Prognostic & 147 & 1,921 \\ 
 & Preventive & 202 & 3,142 \\ \cmidrule(l){2-4} 
 & \textbf{Total} & 617 & 9,133 \\ \midrule
\multirow{4}{*}{OCD} & Factoid & 123 & 1,505 \\ 
 & Diagnostic & \textbf{259} & \textbf{2,307} \\ 
 & Prognostic & 123 & 1,051 \\ 
 & Preventive & 132 & 1,306 \\ \cmidrule(l){2-4} 
 & \textbf{Total} & 637 & 6,169 \\ \bottomrule
\end{tabular}%
}
\caption{MHQA dataset statistics: Distribution across mental health conditions and question types.}
\label{tab:mhqa-stats}
\end{table}

\section{MHQA Dataset}

\subsection{Dataset Overview}
Each instance in the dataset has:
\textbf{Question} which is essentially a 1-2 sentence text.
\textbf{Answer Candidates} where each question has four answer options with single labeled correct option.
\textbf{Domain} to which the question belongs. The domain was categorized into one of the issues, i.e., anxiety, depression, trauma, obsessive and compulsive issues.
\textbf{Question Type} to which the question belongs. The question type was categorized into either factoid, diagnostic, prognostic, or preventive.

MHQA has two parts, namely, (i) MHQA-Gold: a 2,475 human annotated and expert verified QA pair, and (ii) MHQA-B: a \textasciitilde56.1k sized dataset with pseudo labeling based on external knowledge.
A summary of the dataset statistics can be found in table \ref{tab:mhqa-stats}.
\subsection{Data Generation}

\textbf{Data Collection}: We start by collecting PubMed research article abstracts for mental health based on the four chosen domains using specific expert-curated list of keywords for each domain. A comprehensive summary of the keywords can be found in Table \ref{tab:keywords}. We collect abstracts from the years 2000-2024 to ensure the inclusion of sufficiently diverse topics and relevance.
In total, \textasciitilde471k abstracts were collected based on the keywords. The average tokens and words length in the collected abstracts are 243.53 and 206.55 respectively.
\\
\textbf{Generation}: The goal is to generate a meaningful question and corresponding answer using the abstracts as the starting point. Throughout our pipeline, we use the GPT-4o-mini model for LLM generation tasks. We design various criteria for identifying relevant information in the abstract. In summary, the abstract should have (i) mental health information based on one of the four topics, (ii) advanced medical knowledge, and (iii) generalizable concepts and not specific to a particular research. These criteria ensure the questions formed are accurate and relevant. In practice, relevant information can generally be a set of a few sentences centered around a particular topic.

In each instance, the LLM decides if relevant information is present or not. We consider only those abstracts for further processing which are marked as true. 
After topic selection, the model selects the best type of question from the four categories: factoid, diagnostic, prognostic, and preventive, for QA pair generation along with the correct answer candidate. We provide the detailed prompt used for QA generation in \ref{sec:question-type}. Three more plausible yet incorrect candidate options are also generated in reference to the correct answer.
A schematic example of the question generation methodology is also presented in Figure \ref{fig:qa-framework}. Overall, we generate a total of \textasciitilde99.7k initial QA pairs.

At this point, we randomly sample 800 instances from each topic amounting a total of 3,200. We consider these 3,200 QA pairs for further processing into MHQA-Gold.

\subsection{Post hoc Filtering and Validation}
We perform a number of post hoc filtration steps to improve the data quality, and
introduced a validation process for removing inconsistent or non-answerable questions. For example, some question may refer to a related concept from the original abstract which cannot be generalized. We use LLM as a judge (GPT-4o-mini model) to compare the original abstract snippet with the generated question. The LLM decides looking at the information snippet if the generated question is valid, or does it depend on a specific concept from the abstract to be answered, which then is labeled as non-valid. This results in a total of \textasciitilde56.1k from \textasciitilde96.5k forming MHQA-B and 2,475 from 3,200 forming MHQA-Gold initial dataset.
\subsection{Data Annotation}
\textbf{MHQA-Gold}: We employed three trained professional psychologists for data annotation and verification of the MHQA-Gold subset, of which two served as primary annotators, while the third was the secondary annotator. 
The labeling process was carried out independently to ensure anonymity between the annotators about each other's decisions.
Each expert either agrees with the proposed correct option or otherwise presents the correct answer based on their understanding.
We change the option if both experts agree to the same different answer. The secondary annotator acts as a resolver when both the primary annotators have varied opinions, wherein the decision of the secondary annotator would be considered final.
Annotator 1 and annotator 2 show 97.0\% and 97.9\% agreement with the initial MHQA-gold dataset, respectively. We get Cohen’s kappa score of 0.44, which falls within the moderate inter-annotator agreement range.

\textbf{MHQA-B } Due to the large size of the dataset, human validation poses practical challenges for MHQA-B. As a substitute, inspired by the referencing of external sources by the annotators, we curate external medical documents as references. Primarily, we use the NIH and medical examination reference contexts available in MedQuaD \cite{medquad}, MedMCQA \cite{medmcqa}, and DSM-5 Book \cite{dsm5} as validation references for pseudo-labeling. 
For each question from the QA pair, we retrieve the top 3 similar documents from the above sources with a threshold requirement of 0.7 similarity score. If the retrieved documents explicitly contain answerable information about the question, the LLM verifies if the current answer is true or requires change.

\begin{table*}[]
\centering
\resizebox{0.95\textwidth}{!}{%
\begin{tabular}{@{}lcrcrcrcrccc@{}}
\toprule
\textbf{Model} & \textbf{Method} & \multicolumn{2}{c}{\textbf{Factoid}} & \multicolumn{2}{c}{\textbf{Diagnostic}} & \multicolumn{2}{c}{\textbf{Prognostic}} & \multicolumn{2}{c}{\textbf{Preventive}} & \multicolumn{2}{c}{\textbf{Overall}} \\ \midrule
 &  & \multicolumn{1}{c}{Acc.} & F1 & \multicolumn{1}{c}{Acc.} & F1 & \multicolumn{1}{c}{Acc.} & F1 & \multicolumn{1}{c}{Acc.} & F1 & Acc. & F1 \\ \cmidrule(l){3-12} 
\multirow{2}{*}{BERT-base} & base & 29.0 & 28.8 & 27.8 & 27.8 & 34.2 & 34.2 & 31.0 & 31.0 & 30.3 & 30.3 \\
 & SFT & 61.4 & 61.7 & 71.2 & 71.2 & 80.6 & 80.6 & 79.3 & 79.4 & 74.5 & 74.5 \\ \midrule
\multirow{2}{*}{MentalBERT} & base & 22.2 & 22.0 & 23.3 & 23.3 & 25.3 & 25.2 & 24.1 & 24.0 & 23.9 & 23.8 \\
 & SFT & 62.4 & 62.6 & 69.8 & 69.8 & 80.1 & 80.1 & 77.6 & 77.7 & 73.4 & 73.4 \\ \midrule
\multirow{2}{*}{RoBERTa} & base & 28.1 & 28.1 & 31.3 & 31.3 & 28.3 & 28.4 & 28.0 & 28.1 & 29.3 & 29.3 \\
 & SFT & 64.6 & 64.8 & 75.5 & 75.5 & 83.8 & 83.8 & 80.0 & 80.1 &  77.3 & 77.3 \\ \midrule
\multirow{2}{*}{Mental-RoBERTa} & base & 25.3 & 25.3 & 26.7 & 26.7 & 19.4 & 19.3 & 24.1 & 24.1 & 24.1 & 24.1 \\
 & SFT &  63.2 & 63.2 & 72.4 & 72.2 & 79.9 & 79.9 & 79.7 & 79.8 & 75.0 & 75.0 \\ \midrule
\multirow{2}{*}{BioBERT} & base & 30.9 & 30.6 & 34.5 & 34.5 & 30.6 & 30.7 & 34.9 & 35.0 & 33.3 & 33.3 \\
 & SFT &69.3 & 69.4 & 74.4 & 74.4 & 83.8 & 83.8 & \textbf{81.0} & \textbf{81.0} & 77.8 & 77.8 \\ \midrule
Llama-3-8b & zero-shot & \multicolumn{1}{c}{61.8} & 61.8 & \multicolumn{1}{c}{65.8} & 65.6 & \multicolumn{1}{c}{76.6} & 76.4 & \multicolumn{1}{c}{72.7} & 72.6 & 69.7 & 69.6 \\
 & few-shot & 44.8 & 45.7 & 47.7 & 47.7 & 56.7 & 56.6 & 50.9 & 51.1 & 50.3 & 50.4 \\ \midrule
Llama-3.1-8b & zero-shot & \multicolumn{1}{c}{58.3} & 57.9 & \multicolumn{1}{c}{61.6} & 61.5 & \multicolumn{1}{c}{69.0} & 68.6 & \multicolumn{1}{c}{65.2} & 64.7 & 63.9 & 63.6 \\
 & few-shot & 57.9 & \multicolumn{1}{r}{57.9} & 63.2 & \multicolumn{1}{r}{63.2} & 73.2 & \multicolumn{1}{r}{73.2} & \multicolumn{1}{c}{67.4} & 67.3 & \multicolumn{1}{r}{66.0} & \multicolumn{1}{r}{66.0} \\ \midrule
GPT-3.5 & zero-shot & \multicolumn{1}{c}{64.5} & 64.4 & \multicolumn{1}{c}{66.8} & 66.6 & \multicolumn{1}{c}{79.7} & 79.5 & \multicolumn{1}{c}{75.4} & 75.3 & 71.9 & 71.7 \\
 & few-shot & \multicolumn{1}{c}{67.5} & 67.4 & \multicolumn{1}{c}{67.5} & 67.4 & \multicolumn{1}{c}{78.3} & 78.3 & 71.8 & \multicolumn{1}{r}{71.8} & \multicolumn{1}{r}{71.2} & \multicolumn{1}{r}{71.2} \\ \midrule
GPT-4o & zero-shot & \multicolumn{1}{c}{\textbf{74.0}} & \textbf{74.0} & \multicolumn{1}{c}{\textbf{76.6}} & \textbf{76.5} & \multicolumn{1}{c}{\textbf{87.2}} & \textbf{87.2} & \multicolumn{1}{c}{80.6} & 80.6 & \textbf{79.9} & \textbf{79.8} \\
 & few-shot & \multicolumn{1}{c}{72.2} & 72.1 & \multicolumn{1}{c}{75.9} & 75.9 & \multicolumn{1}{c}{85.3} & 85.2 & \multicolumn{1}{c}{79.6} & 79.5 & \multicolumn{1}{r}{78.6} & \multicolumn{1}{r}{78.6} \\ \bottomrule
\end{tabular}%
}
\caption{Benchmarking Results. Performance Comparison of different models and methods on the MHQA-Gold Dataset across four question types and overall, evaluated using accuracy and F1 score}
\label{tab:benchmark-results}
\end{table*}

\begin{figure}
    \centering
    \includegraphics[width=1\linewidth]{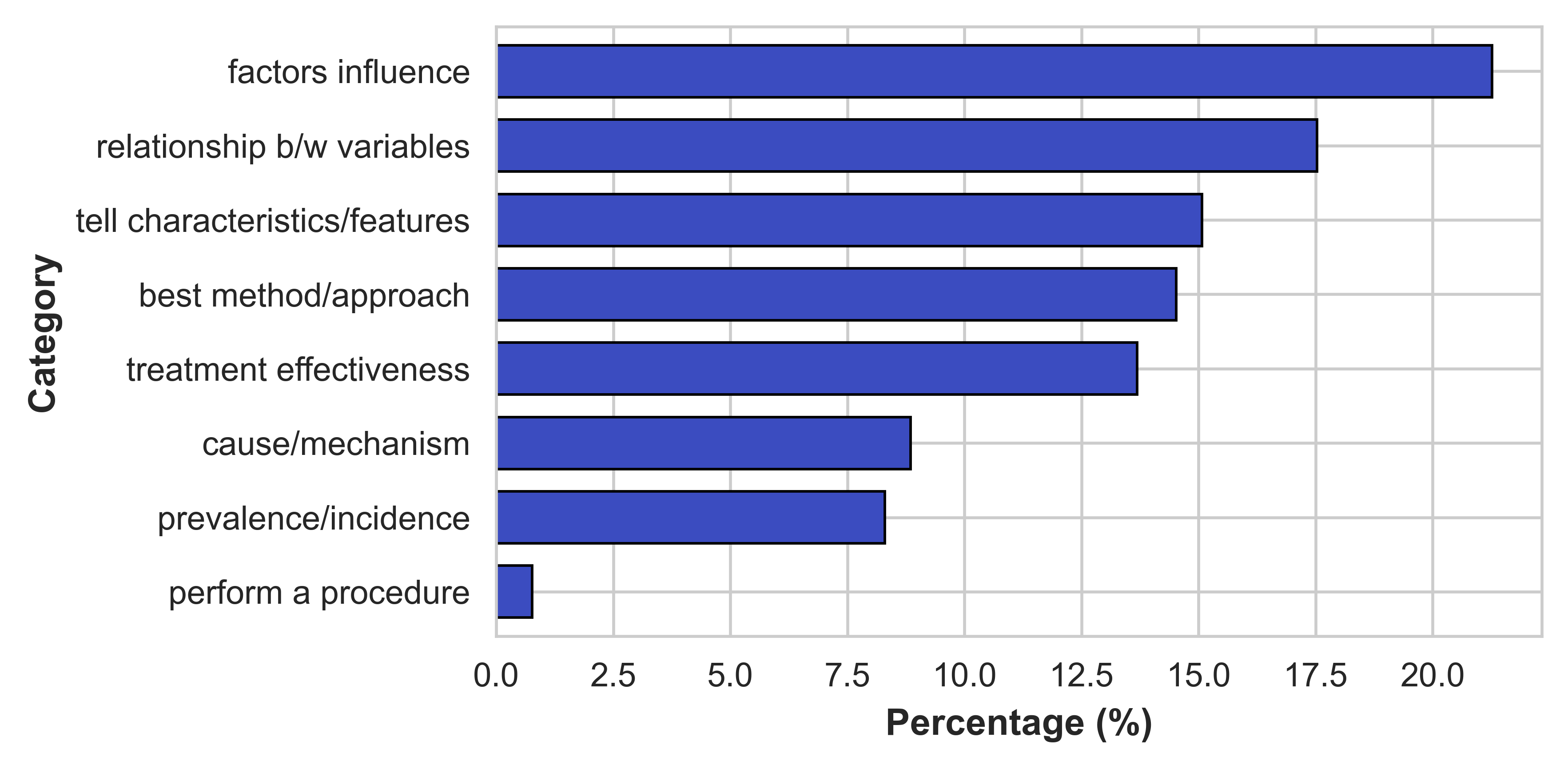}
    \caption{Distribution of different question types in 2K random samples of MHQA dataset.}
    \label{fig:distribution}
\end{figure}

\section{MHQA: A Comparative Analysis}
\textbf{Unique Features: } Our dataset includes four important topics associated with mental health, namely, anxiety, depression, trauma, and obsessive/compulsive issues. This ensures a diversity of topics included for comprehensive analysis of any answering model. 
One of the biggest strengths of MHQA is the presentation of diverse range of questions. 
The various question types hold significance as follows: (i) Factoid-type questions include fact-based verification. Such questions directly test the models' factual knowledge about mental health. (ii) Next, diagnostic questions require LMs to assess symptoms, patterns, and clinical information to determine potential mental health conditions. (iii) While prognostic questions demand the prediction of the course and its subsequent outcomes. Both diagnostic and prognostic types require complex reasoning capabilities to arrive at the correct option. (iv) Finally, the preventive questions type is an important addition for presenting appropriate strategies and interventions required for a particular mental health condition. Overall, our dataset can be seen as a challenging evaluation paradigm for LMs through diverse perspectives. 
\\
\textbf{Rich Knowledge: } Many mental health classification datasets are based on social media platform posts by users. Such text would naturally lack scientific and clinical grounding as well as verification. On the other hand, our dataset includes complex knowledge from research abstracts, which are both medically grounded but also go beyond simple question-answering purview to pose non-trivial and challenging questions. Such knowledge could help inject models with a deeper understanding of mental health. For example, consider Figure \ref{fig:example QA} for examples in MHQA. The questions require strong scientific grounding as well as multi-stage reasoning.
\\
\textbf{Diversity}
Figure \ref{fig:distribution} also shows the distribution of questions in a sample of 2000 instances classified into the most probable question types. Our dataset covers a wide variety of possible medical situations. For example, Many questions are based on 'factors influencing'. A fairly large ratio also focuses on 'relationship between variables' and 'characteristics and features'. Categories like 'cause/mechanism' also suggests the presence of reasoning tasks, which add on to the richness of the dataset.
\section{Experiments}
\subsection{Baseline Models}
We benchmark MHQA-Gold consisting of 2,475 QA pairs on various LMs to leverage their inherent mental health knowledge and reasoning abilities. 
We show the results on the instruct models of Llama-3 8B, Llama-3.1-8B, as well as on GPT-3.5-Turbo, and GPT-4o for Large Language Models. On the other hand, discriminative models include BERT-base \cite{bert}, RoBERTa \cite{roberta}, and BioBERT \cite{biobert}. We also use MentalBERT \cite{ji2021mentalbert} and  MentalRoBERTa \cite{ji2021mentalbert}, which include prior pretraining on mental health data, especially collected from Reddit.

\subsection{Experimental Setup}
The primarily metrics for reporting scores in MHQA-Gold are F1 and Accuracy percentages. We conduct various experimental settings as described here.
\\
\textbf{LLM Prompting } As a general template for LLMs, we prompt the model with the query question, $q$, and the corresponding four options. The model has to strictly choose the output as the correct answer from one of four given options. The chosen option is then matched with the known correct answer along with a brief justification for chosen answer. This is called as zero-shot setup, wherein only the query and instruction is present without additional examples.
\\
\textbf{Few-shot CoT}
We manually design three few-shot Chain of Thought Comparative prompting strategy from the MHQA-B dataset called the few-shot setting.
The prompting strategy follows a three step process for a brief requirement of the question, comparing by justifying or critiquing each option as potential correct answer, and finally presenting the best answer as conclusion.
The LLM then follows similar steps for deducing the correct answer. Three expert curated examples with the steps make up the prompt.
\\
\textbf{Discriminative Models } Along with LLMs, we also experiment with discriminative language models. Some of which have also been trained or previously finetuned on mental health datasets. The models, even though not generative, can perform downstream tasks like choosing best answer from four given options. We perform a direct inference on the models with MHQA-Gold dataset. This helps test the ability of pre-trained versions of the models for efficiency in mental health question answering task.
\\
\textbf{BERT Finetuning} The finetuning approach is similar to \cite{medmcqa} for finetuning discriminative models on multiple choice question answering tasks.
Each option is concatenated with the question, thus creating four such input sequences separated by [SEP] token. 
The encoder generates a separate output token for each of the sequences. We then apply a linear layer with softmax over the four output token. While training data is prepared through this method with known correct option, the correct option while inferencing indicates token with highest softmax. We trained the models for a total of 5 epochs.
\begin{table}[]
\centering
\resizebox{\columnwidth}{!}{%
\begin{tabular}{@{}lrrrrrrrr@{}}
\toprule
\textbf{Model} & \multicolumn{2}{c}{\textbf{Anxiety}} & \multicolumn{2}{c}{\textbf{Depression}} & \multicolumn{2}{c}{\textbf{OCD}} & \multicolumn{2}{c}{\textbf{Trauma}} \\ \midrule
 & \multicolumn{1}{c}{Acc.} & \multicolumn{1}{c}{F1} & \multicolumn{1}{c}{Acc.} & \multicolumn{1}{c}{F1} & \multicolumn{1}{c}{Acc.} & \multicolumn{1}{c}{F1} & \multicolumn{1}{c}{Acc.} & \multicolumn{1}{c}{F1} \\ \midrule
\multicolumn{9}{c}{SFT} \\ \midrule
BERT-base & 72.2 & 72.3 & 76.0 & 76.0 & 72.2 & 72.2 & 77.3 & 77.3 \\
M-BERT & 71.8 & 71.9 & 74.1 & 74.2 & 70.9 & 70.9 & 76.9 & 76.9 \\
RoBERTa& 78.4 & 78.4 & 75.9 & 76.0 & 75.5 & 75.5 & 79.3 & 79.4 \\
M-RoBERTa& 74.9 & 75.0 & 74.2 & 74.2 & 72.8 & 72.8 & 78.0 & 78.1 \\
BioBERT& 77.9 & 77.9 & 77.5 & 77.6 & 75.5 & 75.5 & 80.1 & 80.2 \\ \midrule
\multicolumn{9}{c}{Zero-Shot} \\ \midrule
Llama-3-8b & 70.9 & 70.7 & 66.9 & 66.8 & 67.2 & 67.1 & 73.7 & 73.8 \\
Llama-3.1-8b & 65.2 & 64.8 & 53.2 & 52.5 & 66.1 & 65.9 & 70.7 & 70.6 \\
GPT-3.5 & 70.2 & 70.2 & 71.6 & 71.2 & 69.9 & 69.8 & 75.9 & 75.6 \\
GPT-4o & \textbf{81.0} & \textbf{80.9} & \textbf{80.2} & \textbf{80.2} & \textbf{77.7} & \textbf{77.6} & \textbf{80.6} & \textbf{80.5} \\ \midrule
\multicolumn{9}{c}{Few-Shot CoT} \\ \midrule
Llama-3-8b & 56.6 & 56.6 & 50 & 50.6 & 44.1 & 44.7 & 50.2 & 49.9 \\
Llama-3.1-8b & 66.9 & 66.9 & 65.3 & 65.3 & 63.3 & 63.2 & 68.7 & 68.6 \\
GPT-3.5 & 69.0 & 68.8 & 70.2 & 70.2 & 71.3 & 71.3 & 74.2 & 74.2 \\
GPT-4o & 79.1 & 79.1 & 80.0 & 80.0 & 76.3 & 76.2 & 78.9 & 78.9 \\ \bottomrule
\end{tabular}%
}
\caption{Performance comparison of different models and methods on the MHQA dataset across four topics, evaluated using Accuracy and F1 Score.}
\label{tab:benchmark-results-topic-wise}
\end{table}

\section{Results and Discussions}
\subsection{General}
Table \ref{tab:benchmark-results} summarizes the benchmarking results for various Language Models' performances on the MHQA-Gold dataset. We also compare the performance trends with various question types as well as overall accuracy and F1 scores.
We observe that among various LLMs, GPT-4o shows the overall highest performance with a F1 score of 79.8\% in a zero-shot setting. Similarly, among discriminative models, BioBERT finetuned on MHQA-B dataset ranked highest with a F1 and accuracy score of 81.0\%. 
We observe that both GPT-3.5 and GPT-4o perform better than the smaller parameter models (Llama 3/3.1 8B). This may possibly owe to the larger training base for GPT models. Specifically for our task, the prior Llama 3 8B model performed considerably better than it's later version of Llama 3.1 8B in a zero-shot setting. Llama 3 8b showed a gain of more than 5\% accuracy over Llama 3.1 8b.
BioBERT based SFT model was able to answer QAs more accurately compared to other discriminative models, however RoBERTa performed almost at par with BioBERT with a dip of only 0.5\% accuracy. This is likely attributed to RoBERTa's larger parametric size and extensive pretraining. 

Overall, even after extensive fine-tuning and chain-of-thought based eliminative prompting strategy, the overall score remained below 80\%, and below 70\% for smaller LLMs.
\subsection{Question Type}
Next, we address the question of how does performance look across different tasks.
Interestingly, factoid task has the least scores among the four. Even the best model (RoBERTa SFT) comes with 71.5\% of F1 score. Through this, we understand that factoid type poses as the most challenging variation of MHQA. Subsequently, the models face challenge for answering complex factual based mental health questions.
Among the four, prognostic type showed maximum performance followed by preventive type. We note that both the tasks involves prediction of likely medical outcomes and suggestions. 
Generally, this trend for various question types holds true across all models and settings.
While diagnosis tasks showed lower performance than both, we comment that LLMs have a better grip on mental health outcome prediction and suggestions, while they still lack optimal performance for diagnostic tasks which involves complex reasoning to find correct disorder or related issues.

\begin{table}[]
\centering
\resizebox{0.95\columnwidth}{!}{%
\begin{tabular}{@{}lccc@{}}
\toprule
\textbf{Dataset} & \textbf{Easy} & \textbf{Medium} & \textbf{Hard} \\ \midrule
MedMCQA-psychiatry & 32.1 & 54.7 & 13.1 \\
MHQA & 13.9 & 66.0 & 20.1 \\ \bottomrule
\end{tabular}%
}
\caption{Comparison of question complexity between MedMCQA-psychiatry and MHQA}
\label{tab:mhqa-medmcqa}
\end{table}

\begin{table}[]
\centering
\resizebox{\columnwidth}{!}{%
\begin{tabular}{@{}lcc@{}}
\toprule
\textbf{Model} & \textbf{fs-corr./zs-incorr.} & \textbf{fs-incorr./zs-corr.} \\ \midrule
Llama-3 8B & 238/751 & 713/1723 \\
Llama-3.1 8B & 374/895 & 325/1579 \\
GPT-3.5-Turbo & 252/696 & 270/1779 \\
GPT-4o & 89/499 & 120/1974 \\ \bottomrule
\end{tabular}%
}
\caption{Instances (i) corrected in few-shot (fs-corr.) given incorrect as zero-shot (zs-incorr.), and (ii) incorrect in few-shot (fs-incorr.) given correct as zero shot (zs-corr.).  
Corrected and incorrect answers in the few-shot setting for previously incorrect and correct zero-shot predictions, respectively}
\label{tab:few-shot-analysis}
\end{table}

\subsection{Topic Type}
Table \ref{tab:benchmark-results-topic-wise} shows results for model performances across the four topics. Except for Llama 3.1 8B model, we observe that generally there is a performance dip for obsessive and compulsive disorder category, when compared to other topics. Furthermore, the performance for Anxiety, Depression, and Trauma shows similar trends. This can also provide importance insights into the Language Model's focus areas: while topics like anxiety are also widely studied even with previous datasets, QAs over topics like OCD lack similar performance and are comparatively challenging.
\section{Analysis}
\subsection{CoT few-shot instruction tuning}
We observed that for Llama 3.1 8B model, CoT setting showed an increase of 2.4\% in F1 score and 2.1\% in accuracy. The trend generally holds true for all question types, except for factoid where accuracy slightly decreases. GPT-3.5 showed an overall slight decrease of 0.5\% F1 score with few shot setting. On a closer analysis, we find that GPT-3.5 leverages the setting for both factoid and diagnostic type. On the other hand, GPT-4 slightly underperformed with this setting. Table \ref{tab:benchmark-results-topic-wise} also shows a similar trend across various topics, where GPT-4 slightly under performs with few-shot setting while Llama-3.1 8B outperforms its counterpart zero-shot setting.
Llama-3 8B showed significant dip in performance after few-shot examples. On a qualitative analysis of the responses, we find that an advanced CoT is difficult for Llama-3 8B to follow. However, considering Llama 3.1 8B, few shot CoT helps enhance its performance.

Table \ref{tab:few-shot-analysis} interestingly shows the number of questions correctly answered with few-shot given they were previously answered incorrectly in zero-shot setting. We then also show the question incorrectly answered with few shot given they were previously correctly answered in zero-shot setting. We observe that for larger models like GPT, it performs at almost par with zero-shot settings, although it can  adversely affect its natural answering ability. This can potentially be attributed to the limited examples occasionally inducing bias by generalizing the methodology to answer a given question.

\subsection{Supervised Finetuning}
We finetune the pretrained discriminative models in a supervised fashion over the MHQA-B dataset containing similar distribution. Finetuning significantly improves the model performances when compared to it's pretrained version. This is in conjunction with the fact that both MHQA-B and MHQA-Gold (used for inference results) follow similar distribution, and the models are greatly benefited by the knowledge incorporation, \textit{i.e.}, the models are able to learn mental health related knowledge.
The trend of significant improvement in performance after SFT is similarly also confirmed by previous studies like \cite{medqa, pubmedqa, angst}.
BERT-SFT performs slightly better than Mental BERT-SFT, and similarly RoBERTa-SFT is superior to MentalRoBERTa-SFT. Overall, all these models show a minimum of 40\% increase in F1 scores after finetuning. The results for Mental-RoBERTa and Mental-BERT are also noteworthy. Despite being trained on Reddit and similar mental health corpora, these variants do not seem to incorporate the clinical knowledge necessary for MHQA. Consequently, both the base and SFT models of BERT and RoBERTa outperform their mental health-trained counterparts by 1-2\%.
\subsection{Comparison with other dataset}
While other mental health QA related datasets are either long form generation task or non-English, MedMCQA \cite{medmcqa} consisting of Psychiatry questions comes close to MHQA in pattern.
We compare the questions present in our dataset with those in MedMCQA-Psychiatry by randomly sampling 800 questions from both datasets. 
Further, an LLM (GPT-4) classifies them into three labels: (1) Easy - Requiring direct knowledge for answering; (2) Medium - Requires reasoning over multiple topics; (3) Hard - Requires complex reasoning and synthesis of multiple domains.
Table \ref{tab:mhqa-medmcqa} shows that our dataset follows a fair mixture of all three question types, with Medium and Hard questions percentage exceeding MedMCQA.

\section{Conclusion}
Our work proposes a novel multiple choice questions format-based mental health question answering dataset, called MHQA, for evaluating the performance of language models with diverse question and topic types. Our benchmarking results highlight the superior performance of GPT-4o as an LLM. However, other models still significantly underperform, highlighting challenges in handling complex mental health reasoning. 
We also find that discriminative models show great capability to learn in a supervised finetuning fashion, sometimes even outperforming GPT-4o. It was also observed that Factoid QA is particularly challenging compared to other reasoning and predictive tasks. 



\section*{Limitations and Future Work}
Our work still poses limitations which also serve as scope for future work. The pseudo-labeling process does not validate each pair but only those with 0.7 or higher similarity scores. Further the actual correction still depends on the relevant data present. Our current topics are also limited to four domains, while there is a scope for scaling. Additional methods to improve scores like variable prompting methods, RAG, Knowledge Graphs, etc. can also be used.

\section*{Ethics Statement}
The mental health dataset may contain disturbing terms associated with issues like depression and anxiety.
\section*{Acknowledgements}
Suraj Racha is funded and supported by the Prime Minister Research Fellowship (PMRF) for his Ph.D. work. This research is partially supported by BharatGen.

\bibliography{custom}

\newpage
\appendix

\section{Appendix}
\label{sec:appendix}
\section{Data Collection}
The specific keywords used for data collection can be found in table \ref{tab:keywords}.

\begin{table}[h]
\centering
\resizebox{\columnwidth}{!}{%
\begin{tabular}{@{}>{\raggedright}p{0.4\columnwidth} >{\raggedright\arraybackslash}p{0.6\columnwidth}@{}}
\toprule
\textbf{Category} & \textbf{Keywords} \\ \midrule
Anxiety & anxiety, anxious \\
Depression & depression, depressive, melancholia, dysthymia \\
Trauma & PTSD, post-traumatic stress, post-traumatic stress disorder, post traumatic stress disorder, Bipolar disorder \\
OCD & OCD, obsession, obsessive \\ \bottomrule
\end{tabular}%
}
\caption{Keywords used for abstracts collection for each category}
\label{tab:keywords}
\end{table}



\section{Prompts used}

\onecolumn
\subsection{Question Generation}
\label{sec:question-type}
\hrule
\vspace{-3mm}
\ttfamily
\begingroup
\begin{longtable}{p{0.95\textwidth}}
\endfirsthead
\endhead
\endfoot

\textbf{Role:} You are an advanced language model tasked with generating questions from research abstracts in the field of mental health. Your questions will cover factoid, reasoning, and yes/no types, derived from details within each abstract.\\
\vspace{2pt}
\textbf{Goal:} Create a single, insightful question from each abstract, incorporating relevant details. This question will help users engage with the content and aid in building a knowledge graph. \\ 
\vspace{1pt}
\textbf{Task Instructions:}
\begin{itemize}
    \item \textbf{Identify Question Type:}
    \begin{itemize}
        \setlength{\itemsep}{0pt}
        \setlength{\parskip}{0pt}
        \item Determine if a question can be derived from the abstract.
        \item Choose the best question type:
        \begin{itemize}
            \item \textbf{Factoid:} A question with a concrete answer. Structure as a multiple-choice question (MCQ) with one correct answer and three distractors.
            \item \textbf{Prognostic:} A question that predicts future outcomes or developments based on the abstract's content.
            \item \textbf{Diagnostic:} A question that identifies the cause, origin, or underlying factors of a problem or condition.
            \item \textbf{Preventive:} A question that addresses strategies, methods, or interventions to avoid, reduce, or mitigate a problem.
        \end{itemize}
    \end{itemize}

    \item \textbf{Question Construction:}
    \begin{itemize}
        \setlength{\itemsep}{0pt}
        \setlength{\parskip}{0pt}
        \item The question topic should strictly be in these four categories: \texttt{Trauma, Anxiety, Depression, Obsessive/Compulsive Disorders}. If the question does not fit any of these topics, mark it as \texttt{false}.
        \item Questions should be related to health or medical topics.
        \item Select statements from the abstract that contain essential information but hide the actual answer.
        \item Aim for \textbf{advanced medical graduate-level questions}.
        \item Ensure that the question represents a medical or health-related concept.
        \begin{itemize}
            \item Example: \texttt{"Which three ferroptosis-related genes were identified as part of the diagnostic signature for major depressive disorder?"}
            \item Example: \texttt{"Which selective serotonin reuptake inhibitor is noted as a first-line treatment for monopolar depression?"}
        \end{itemize}
        \item Ensure the question is \textbf{not specific} to a research paper but captures broader concepts.
        \item \textbf{Avoid} referencing specific studies (e.g., \texttt{"According to the study," "The research suggests," etc.}).
        \item \textbf{Avoid numerical/percentage-based questions}.
    \end{itemize}

    \item \textbf{Categories and Edge Cases:}
    \begin{itemize}
        \item Focus on \textbf{mental health or related medical topics}.
        \item If factoid questions are not suitable, consider \textbf{prognostic, diagnostic, or preventive} types.
        \item Rephrase ambiguous or complex terms for clarity while maintaining scientific accuracy.
    \end{itemize}
\end{itemize}

\textbf{Output Format:} \\
\begin{itemize}[label=-]
    \setlength{\itemsep}{0pt}
    \setlength{\parskip}{0pt}
    \item \textbf{Id:} Research paper ID.
    \item \textbf{Valid Question:} \texttt{True} if a valid question can be generated, otherwise \texttt{False}.
    \item \textbf{Question Topic:} \texttt{Trauma, Anxiety, Depression, Obsessive/Compulsive Disorders}.
    \item \textbf{Type of Question:} \texttt{Factoid, Prognostic, Diagnostic, Preventive}.
    \item \textbf{Question:} The generated question.
    \item \textbf{Options:} Four answer choices in \texttt{< >}, with one correct and three distractors.
    \item \textbf{Correct Answer:} The correct choice.
\end{itemize}

Here are the abstracts of the research papers. Read each abstract and generate a question following the instructions. \\

\texttt{\{abstracts\}} \\

\textbf{Abstract Format:}
\begin{verbatim}
Id: {pmid}
Title: {title}
Abstract: {abstract}
\end{verbatim}
\end{longtable}
\endgroup
\vspace{-23pt}
\hrule

\normalfont

\onecolumn
\subsection{Question Validation}
\hrule
\ttfamily
\begingroup
\begin{longtable}{p{0.95\textwidth}}
\endfirsthead
\endhead
\endfoot
You are a medical expert. You are given an abstract from a research paper and a question formed based on the abstract. Your goal is to determine whether the question is valid based on the following criteria. \\
\vspace{1pt}
\textbf{Conditions for validity of a question given a abstract:} \vspace{-5pt}
\begin{itemize}[label=-]
    \setlength{\itemsep}{0pt}
    \setlength{\parskip}{0pt}
    \item It directly aligns with the information in the abstract meaning that the question should be answerable using the exact snippet from the abstract.
    \item The information in the snippet represents a generally applicable finding meaning that if the snippet presents a conclusion or insight that extends beyond the specific study (i.e., it is a widely applicable or generalizable concept), the question is valid.
    \item It does not introduce ambiguity or incompleteness meaning that if the snippet presents very specific or preliminary findings and the question generalizes or omits key details, making the answer incomplete or misleading, the question is not valid.
    \item The question has a uniquely distinct answer based on the abstract meaning that if multiple interpretations exist, or if the study only proposes a possibility rather than confirming a fact, then the question is not valid.
\end{itemize}

\textbf{Task Instructions:} \vspace{-5pt}
\begin{itemize}
    \setlength{\itemsep}{0pt}
    \setlength{\parskip}{0pt}
    \item You'll be given multiple abstracts and questions.
    \item Read each abstract and question carefully.
    \item For each abstract and question pair do the following:
    \begin{itemize}
        \setlength{\itemsep}{0pt}
        \setlength{\parskip}{0pt}
        \item Identify the exact snippet in the abstract that the question is based on.
        \item Check each condition against the question and the abstract to determine the validity of the question.
    \end{itemize}
\end{itemize}

\textbf{Learn from these examples:} \\
\vspace{0.5pt}
\textbf{Example 1:} \\
\textbf{ID:} 1 \\
\textbf{Question:} How does the dual experience of parental incarceration and foster care impact youth mental health outcomes ? \\
\textbf{Abstract:} While there are various pathways by which children experience parental incarceration or foster care, involvement in either system is associated with adverse health outcomes. Despite co-occurring risk factors for parental incarceration and foster care, little is known about the prevalence or characteristics of youth navigating both of these experiences. This study details the prevalence of youth at the intersection of parental incarceration and foster care, their demographic characteristics, and heterogeneity in their mental health. Data come from the 2019 Minnesota Student Survey with 112,157 eighth-, ninth-, and eleventh-grade students. Logistic regression with interactions between parental incarceration and foster care predict associated odds of youth's anxiety and depression; self-injurious behavior, suicidal ideation and attempt; and mental health diagnoses and treatment. Nearly 2000 of students experienced both parental incarceration and foster care, with a disproportionate number of those identifying as youth of color, experiencing poverty, and/or living in rural communities. Both parental incarceration and foster care were separately linked with poor mental health, yet experiencing both was associated with higher odds of anxiety, depression, self-injury, suicidal ideation, suicide attempt, diagnosis, and treatment. Youth with proximal multiplicative exposure (recent foster care and current parental incarceration) reported the most adverse mental health symptoms. The study emboldens what is known about the inequitable distribution of parental incarceration and foster care. These findings highlight the association between dual-systems-impacted youth and mental health indicators, with important implications for increasing access to mental health services while simultaneously calling for systems change. \\
\textbf{Snippet:} Both parental incarceration and foster care were separately linked with poor mental health, yet experiencing both was associated with higher odds of anxiety, depression, self-injury, suicidal ideation, suicide attempt, diagnosis, and treatment. Youth with proximal multiplicative exposure (recent foster care and current parental incarceration) reported the most adverse mental health symptoms. \\
\textbf{Justification: } The question is valid because the information can be commonly applied as asked in the question. The study establishes a clear relationship between the dual experience and mental health outcomes, making the question answerable without ambiguity.\\
\textbf{Decision:} Valid \\
\vspace{0.5pt}

\textbf{Example 2:} \\
\textbf{ID:} 2 \\
\textbf{Question:} What is noted as a potential biomarker for treatment response in major depressive disorder? \\
\textbf{Abstract:} Repetitive transcranial magnetic stimulation (rTMS) is an effective and safe treatment for major depressive disorder (MDD). rTMS is in need of a reliable biomarker of treatment response. High frequency (HF) dorsolateral prefrontal cortex (DLPFC) rTMS has been reported to induce significant changes in the cardiac activity of MDD patients. Low frequency DLPFC rTMS has many advantages over HF-DLPFC rTMS and thus this study aims to further investigate the effect of low frequency 1000Hz right hemisphere (R)-DLPFC rTMS on the cardiac activity of MDD patients, as well as the potential of using electrocardiogram (ECG) parameters as biomarkers of treatment outcome. Baseline ECG sessions were performed for 19 MDD patients. All patients then underwent 40 sessions of accelerated 1000Hz R-DLPFC rTMS one week after the baseline session. Heart rate (HR) significantly decreased from the resting period to the first and third minute of the 1000Hz R-DLPFC rTMS period. Resting HR was found to have a significant negative association with treatment outcome. Prior to Bonferroni correction, HR during stimulation and the degree of rTMS-induced HR reduction were significantly negatively associated with treatment outcome. No significant changes were observed for the heart rate variability (HRV) parameters. Sample size (n=19); the use of electroencephalography equipment for ECG; lack of respiration monitoring; relatively short recording duration for HRV parameters. This novel study provides further preliminary evidence that ECG may be utilized as a biomarker of rTMS treatment response in MDD. ClinicalTrials.gov Identifier: NCT04376697.
Snippet: Relatively short recording duration for HRV parameters. This novel study provides further preliminary evidence that ECG may be utilized as a biomarker of rTMS treatment response in MDD. \\
\textbf{Justification:} The question is not valid because the study proposes ECG as a potential biomarker, rather than confirming it as an established fact. The phrasing of the question suggests a definitive answer, but the study only presents preliminary evidence. Since the snippet does not confirm ECG as a widely accepted biomarker, the question lacks clarity and is misleading. \\
\textbf{Decision:} Not Valid \\
\vspace{0.5pt}

\textbf{Output format for each question:} \\
\textbf{ID:} <id> \\
\textbf{Snippet:} <a snippet from the given abstract> \\
\textbf{Justification:} <justification for your decision> \\
\textbf{Decision:} <"Valid" or "Not Valid"> \\
\vspace{0.5pt}
Read the following questions and abstracts and determine the validity of each question:
\{question\_abstract\_pairs\} \\
\textbf{Output:} \\
\textbf{ID:} \\
\end{longtable}
\endgroup

\hrule

\vspace{3mm}

\normalfont

\subsection{Benchmarking Prompt}
\vspace{-5pt}
\begin{table}[H]
\centering
\renewcommand{\arraystretch}{1.3} 
\noindent\begin{adjustbox}{max width=\textwidth}
\ttfamily 
\begin{tabular}{p{\textwidth}}
\hline
\textbf{Role:} You are a highly skilled medical expert tasked with evaluating multiple-choice questions. Your role is to select the single most accurate and contextually correct option from the given choices. Do not favor any option based on its order. Evaluate all options carefully and justify your choice implicitly. \\

\textbf{Output Format:} \\
\texttt{Correct Option:} <1 or 2 or 3 or 4> \\
\texttt{Justification:} <justification for the answer> \\

\textbf{Question:} \{question\} \\
\textbf{Options:} \\
\texttt{Option1: \{op1\}} \\
\texttt{Option2: \{op2\}} \\
\texttt{Option3: \{op3\}} \\
\texttt{Option4: \{op4\}} \\

\textbf{Respond with the letter corresponding to your choice (1 or 2 or 3 or 4) followed by a justification.} \\
\texttt{Correct Option:}  \\

\hline
\end{tabular}
\end{adjustbox}
\end{table}

\normalfont

\subsection{Pseudo Labelling}
\vspace{3mm}

\section{Human Annotation}
Criteria for annotations
\\
More details

This is a section in the appendix.
\newpage
\section{Examples}
Table \ref{tab:example-table} shows more examples of various questions available in the MHQA dataset.
\begin{table*}[t]
\centering
\resizebox{0.9\textwidth}{!}{%
\begin{tabular}{@{}p{0.8\textwidth} p{0.15\textwidth}@{}}
\toprule
\textbf{Example Questions} & \textbf{Brief} \\ \midrule
\multicolumn{2}{@{}l}{\textbf{Factoid}} \\ \midrule
Which receptor activation has been shown to effectively counter opioid-induced respiratory depression? & Fact verification type \\
Which compound has shown efficacy as an augmentation treatment for serotonin reuptake inhibitor refractory obsessive-compulsive disorder? & Fact verification type \\
Which anxiety disorder was noted for having the most prevalence of negative beliefs about the uncontrollability and danger of thoughts? & Fact verification type \\ \midrule
\multicolumn{2}{@{}l}{\textbf{Diagnostic}} \\ \midrule
What underlying factor may contribute to poorer rehabilitation outcomes in individuals with spinal cord injury who have psychological need? & Diagnosing the disease \\
Which psychological trait is associated with impaired memory for emotional stimuli in individuals? & Diagnosing the disease \\
What cognitive impairment is associated with excessive obsessive-compulsive symptoms, impacting an individual's ability to predict the present based on past experiences? & Diagnosing the disease \\ \midrule
\multicolumn{2}{@{}l}{\textbf{Prognostic}} \\ \midrule
How might lockdown-induced changes in sleep duration and disturbances affect the long-term mental health of children and adolescents with ADHD? & Predicting the outcome \\
What long-term effects on maternal mental wellbeing might result from continuous lockdowns during the COVID-19 pandemic? & Predicting the outcome \\
How does cognitive reserve interact with sleep onset/maintenance difficulties to affect cognitive performance in older adults? & Predicting the outcome \\ \midrule
\multicolumn{2}{@{}l}{\textbf{Preventive}} \\ \midrule
What could be considered an effective treatment option for patients experiencing visual complaints associated with HPPD and anxiety symptoms? & Provide or predict solution \\
Which factors should be considered for preventing depression and anxiety in women post-acute myocardial infarction? & Provide or predict solution \\
What factors should clinicians consider when addressing selective eating in children with anxiety/obsessive spectrum disorders? & Provide or predict solution \\ \bottomrule
\end{tabular}%
}
\caption{Example questions categorized by type and their corresponding explanation.}
\label{tab:example-table}
\end{table*}
\end{document}